\documentstyle[twocolumn,lingmacros]{article}
\newtheorem{definition}{Definition}
\begin{document}
\title{{\bf Semantics of Complex Sentences in Japanese}}

\author{Hiroshi Nakagawa \and Shin-ichiro Nishizawa}

\date{Dept. of Electronics and Computer Engineering, Yokohama National
University\protect\\
e-mail: $\{$ nakagawa, shin $\}$@naklab.dnj.ynu.ac.jp}
\maketitle

\begin{abstract}
The important part of semantics of complex sentence is captured as
relations among semantic roles in subordinate and main clause
respectively.  However if there can be relations between every pair of
semantic roles, the amount of computation to identify the relations that
hold in the given sentence is extremely large. In this paper, for
semantics of Japanese complex sentence, we introduce new pragmatic roles
called {\it observer} and {\it motivated} respectively to bridge
semantic roles of subordinate and those of main clauses. By these new
roles constraints on the relations among semantic/pragmatic roles are
known to be almost local within subordinate or main clause. In other
words, as for the semantics of the whole complex sentence, the only role
we should deal with is a {\it motivated}.
\end{abstract}

\section{Introduction}

Our aim is to formalize constraints that are needed to develop a
parser based on unification grammar (called ``UG'' henceforth) so that
our parser can deal with variety of types of sentences in Japanese.
However just parsing syntactically is not enough for natural language
understanding.  One important and necessary task to be done, when a
parser processes a discourse in Japanese, is the so called zero
anaphora resolution.  All of syntactic, semantic, and pragmatic
constraints are to be involved to resolve zero anaphora.  Of course, some
of omitted pronouns are syntactically resolved. For instance, VP with
suffix {\it te} is not regarded as a clause but a conjunct VP.
Therefore the subject of the VP with {\it te}, which is possibly
omitted from surface, should corefer with the subject of the sentence.
One example is

\enumsentence{\label{te}
\begin{tabular}{llll}
Hanako & -wa & $\phi_{1subj}$ & samuku-te \\
    & -TOPIC & & feel cold\\
\end{tabular}
\begin{tabular}{llll}
$\phi_{2subj}$ & mado & - o & sime-ta.\\
 & window & - ACC & closed.\\
\end{tabular}\\
`Hanako felt  cold and closed the window.'}

where both of zero subjects $\phi_{1subj}$ and $\phi_{2subj}$
\footnote{
Henceforth, $\phi_{\$\$\$}$ means zero \$\$\$.., where \$\$\$.. is either
grammatical, semantic or pragmatic role. For instance, $\phi_{subj}$ means
zero subject,$\phi_{agt}$ means zero {\it agent}, $\phi_{exp}$ means zero
{\it experiencer}, and so forth.}
 refer to the sentential topic {\it
Hanako}
\footnote{ `Hanako' is a typical girl's name.}
.  In this example, one of the possible accounts for this
interpretation is the following. Zero subject of {\it -te} phrase is [ +
anaphoric, + pronominal ] or PRO in GB term \cite{Sells85}. As
the result, $\phi_{1subj}$ is controlled by the subject $\phi_{2subj}$
of the main VP, which is also zero subject.  $\phi_{2subj}$ is, in GB
term, [ - anaphoric, + pronominal ] or pro.  The sentential
topic Hanako is the only possible antecedent of this zero
subject in this example.  However, in complex sentences, things are
quite different.  Consider the following sentence.

\enumsentence{\label{samu0}
\begin{tabular}{llll}
Hanako & - wa & [ $\phi_{1subj}$ & samu \\
  & - TOPIC & [ & feeling cold \\
\end{tabular}
\begin{tabular}{lr}
 -gat-ta & node]\\
 behaved like & because] \\
\end{tabular}
\begin{tabular}{llll}
$\phi_{2subj}$ & mado - o &  sime-te & yat-ta.\\
 & window - ACC & close & gave.\\
\end{tabular}\\
1. `Since Hanako behaved like feeling cold, I closed the window.'\\
2. `Since I behaved like feeling cold, Hanako closed the window.'}

If contextually we can take only Hanako and the speaker of this sentence
as candidates of antecedent of $\phi_{1subj}$ or $\phi_{2subj}$,
intuitively the following two interpretations are equally likely.

\begin{description}
\item[a.] $\phi_{1subj}$ = Hanako, $\phi_{2subj}$ = speaker
\item[b.] $\phi_{1subj}$ = speaker, $\phi_{2subj}$ = Hanako
\end{description}

Therefore $\phi_{1subj}$ and $\phi_{2subj}$ are both pro. In fact this
fact is well known among Japanese linguists, i.e.
\cite{Sells85,takubo}.  As a result, zero anaphora resolution of complex
sentence is not only to be done syntactically, but also to be done
pragmatically and/or semantically.  One of the promising candidate for
this is the centering theory \cite{Brennan,LynMasayo}.  To apply the
centering theory that is originally for a sequence of sentences, namely
{\it discourse}, we regard the subordinate clause and the main clause as
a segment of discourse respectively. Moreover Hanako who is marked by
`wa' is regarded as the topic for these two clauses. Then, the topic
Hanako is the strongest candidate for the backward center of the
subordinate clause. Therefore the backward center of the subordinate
clause is Hanako, and consequently zero subject $\phi_{1subj}$ refers to
Hanako. By the same way as the subordinate clause case is dealt with,
the zero subject of the main clause $\phi_{2subj}$ is known to refer to
Hanako, too. This result is neither interpretation {\bf a} nor {\bf b}
shown above. Another candidate is the property sharing thoery
\cite{Megumi88}. In her theory, since the both of zero subjects share
the subjecthood, both of them finally are known to refer to Hanako that
is the topic for both of these clauses. Therefore the property sharing
theory also fails to account for the intuitive interpretations.

Then we shift our attention to more microscopic one, in
which ,roughly speaking, the important part of semantics of complex
sentence is formalized as relations among semantic roles that appear in
the main clause or the subordinate clause. At the first glance, the
constraints about these relations are not local in terms of main or
subordinate clauses. In other words, semantic roles that appear in
subordinate clause and semantic roles that appear in the main clause
seem to be directly constrained by the constraints of complex sentence.
However, looking more carefully, we find that the constraints of
subordinate clause and the constraints of main clause are represented as
local constraints by introducing the new notion of {\it motivated} which
is characterized as a person who has enough reason to act as the main
clause describes.  More precisely, {\it motivated} is one of the
pragmatic roles that appear in a subordinate clause, and the constraints
in subordinate clause are stated as identity relations between {\it
motivated} and other semantic/pragmatic roles appearing in subordinate
clause.  Therefore these constraints are local in subordinate clause.
The constraints in main clause are stated as identity relations between
{\it motivated} which comes from subordinate clause, and other semantic
roles appearing in main clause.  Therefore in understanding the main
clause we don't have to be care about semantic/pragmatic roles in
subordinate clause other than a {\it motivated}. In this sense, the
constraints in the main clause can be treated as almost local
constraints of the main clause.

The next question is how to represent the semantics of complex sentence
in feature structure( called FS henceforth ).  For this, we should write
down the constraints about these relations among semantic/pragmatic
roles in a feature structure formalism.
Due to the space limitation, in this
paper we mainly pursue the constraints about semantic feature
structures.

\section{Hierarchical Structure of Complex Sentence}

We pay our attention to the general structure of Japanese utterance
which is helpful to represent semantics of complex sentence.  Several
Japanese linguists have already proposed the general structure of
Japanese utterances \cite{mikami,minami,takubo,gunji89}. Mikami
categorized clauses into three classes, namely `open', `semi-open' and
`closed.' This categorization indicates how freely the content of
clause interacts with the outside of clause. For instance, they are
categorized by the degree of possibilities of coreference between zero
pronouns inside the subordinate clause and nominal or topic that
appear in the main clause. Following Mikami's idea, Minami proposed
four levels, namely level A, B, C and D which correspond roughly to
VP, proposition, sentence without communication mood and utterance
which takes into account a hearer, respectively.
\cite{takubo} divided level A into two levels.
One of them corresponds to VP, the other corresponds to VP + a certain
kind of subject which is called ``objective subject.''
Gunji proposed the more detailed structure, in which starting from
predicate, say, verb and adjective, objects, voice,
subject, aspect, tense, modality, topic and mood are or
might be sequentially added to make an informationally more fulfilled
sentence component. Finally, it ends up with an utterance. In Gunji's
structure, some node can have more than two daughter nodes to make
more complex sentence. Following them, the structure of the so called
(cluase level) complex sentence is the following shown in
Fig.1.

In Fig.1 , Sub-Clause and Conjunct mean subordinate clause and
conjunctive particle respectively.  Note that Fig.1 represents not only
the hierarchical structure but also the word order of a complex sentence
in Japanese. The structure is almost the same as Gunji's structure
except for explicitly showing complex proposition, subordinate-clause
and conjunctive-particle that are newly added to deal with complex
sentences. Note that `Comment' appearing in `Sub-Clause' has the same
structure as `Comment' appearing just below `Judgement'. That is to say,
`Comment' is recursively defined. However, in practice, the more the
level of depth of recursively appearing `Comment' is, the less
comprehensible the sentence is.

\begin{verbatim}

                    Utterance
                     /   \
             Judgement    Mood
              /    \
           Topic   Comment
                   /      \
              Event        Modal
             /     \
   Sub-Clause       Proposition
   /        \        /     \
Comment  Conjunct  Process  Tense
                  /       \
        Action/State      Aspect
        /          \
   Subject          VP
      /  \
 Object   VP
         /  \
Predicate    Voice

\end{verbatim}
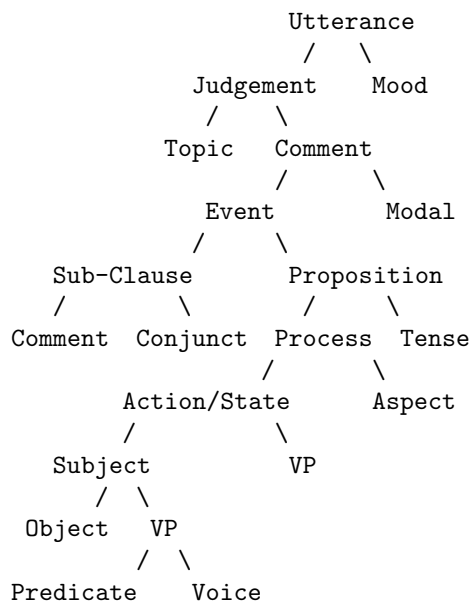
Figure.1: The hierarchical structure of Japanese utterances\\

\section{Subordinate Clause}

In this section, at first we show the predicate categories used in the
subordinate clauses that we deal with in this paper, in
Table.\ref{Table1}.
\begin{table}[b]
\begin{tabular}{|l|l|}\hline
1 \label{ns} & non-subjective predicate\\
\hline
2 \label{sv} & subjective verb\\
\hline
3 \label{sa} & subjective adjective \\ & without verbal suffix {\it garu}.\\
\hline
4 \label{sag} & subjective adjective \\ & with verbal suffix {\it garu}.\\
\hline
5 \label{satg} & {\it verb} + {\it ta-garu}\\ & (behave as s/he wants to ``{\it
verb}'')\\
\hline
6 \label{p} & transitive passive and intransitive passive \\ & (adversity
passive).\\
\hline
\end{tabular}
\vspace*{-2mm}
\caption[Table1]{Predicate Categories}\label{Table1}
\end{table}
In each category of 2,3,4,5 and 6,
exists there a person who is affected by the situation described by
the subordinate clause.  On the contrary, in category 1, there is
not necessarily an explicit affected person. In our theory, this
affected person plays a key role for semantics of complex sentence. As
the result, in general we cannot derive a useful result for category 1
 in our theory.  Therefore we don't deal with category 1
in this paper.

At this moment, we should explain the nature of the so called subjective
predicate mentioned in Table.\ref{Table1}.  In short a subjective
predicate describes the {\it experiencer}'s inner state which can
exclusively be known by the {\it experiencer} him/herself.

Next we focus on verbal suffix {\it garu}. Firstly we show {\it
garu}'s syntax. {\it Garu} is the present form and its root form is
{\it gar}. Therefore inflections are as follows: {\it gar-re},{\it
gar-i}, etc. In addition, {\it garu} has an allophonic
root form {\it gat} and, {\it gat-ta}(past-form), {\it
gat-teiru}(progressive-form) and so on are derived from {\it gat}.
Some of these forms will appear in our examples.  Next we talk about
the semantics of {\it garu}.  {\it Garu} roughly means ``show a sign
of'' or ``behave like ..ing''\cite{Ohye}.  Also in \cite{Palmer} its
semantics is informally explained, however our proposal is to
formalize {\it garu}'s semantics in UG or more generally in
computational linguistics.  For this, first of all, we introduce a new
pragmatic role called {\it observer}.

\begin{definition}[Observer]
Observer is a person who directly observes or is indirectly informed the
situation described by the proposition part. Therefore an observer has a
certain evidence to be convinced that that situation actually happens.\\
\end{definition}

Although this notion of {\it observer} shares a large part with PIVOT of
\cite{Iida}, our notion of {\it observer} is introduced only by {\it
garu}.  Therefore it is much narrower notion.  As you will see later,
this newly introduced role is playing a key role which bridges semantic
roles of subordinate clause to semantic roles of main clause.

As for an {\it observer} introduced by {\it garu}, one of the widely
known consequence about the nature of subjective predicate is the
following. In a sentence, if a subjective adjective is used without
being followed by a verbal suffix
{\it garu}, the {\it experiencer} of the subjective adjective should be the
speaker of the sentence.

The next thing we should do about a newly introduced notion of {\it
observer} is to make clear the way to deal with it in FS. First of all,
in our FS, a semantic content:SEM is basically a soa (state of affair)
form of situation semantics. However we use semantic role like ``{\it
agent}'', ``{\it patient}'', ``{\it experiencer}'', and so on, as
argument roles of soa.  Since an {\it observer} observes the situation
which is characterized by a soa, if we know that there exists an {\it
observer}, the observed soa is embedded in observing situation, which,
in turn, is embedded in the whole semantic content. In this sense, the
observed soa's argument role is {\it observed}.  But as far as we have
no confusion, we omit role name `observed' henceforth.  A typical schema
of SEM of FS of this type is the following. Note that we use {\it garu}
as a value of the relation feature meant by `rel.' The English gross of
this relation {\it garu} is `observe.'

\enumsentence{
SEM =
\outerfs{rel: {\it garu}\\
         observer:$\:$\fbox{o}\\
         soa:\outerfs{rel:R\\agent:$\:$\fbox{a}\\experiencer:$\:$\fbox{e}\\
                             patient:$\:$\fbox{p}\\....\\}\\}}

Now we explain the semantics of clause which consists of subjective
adjective with {\it garu} or {\it ta-garu}, that are in categories
4 and 5.
These categories' forms are ``$\phi_{exp}$ P-garu'' or its past form
``$\phi_{exp}$ P-gat-ta'', where P is a subjective adjective (category
4 in Table.\ref{Table1}) or is a verb followed by {\it ta-gar} (category
5 in Table.\ref{Table1}), and $\phi_{exp}$ is the {\it experiencer} of P which
is possibly zero. In these categories, there exist observers who are
not the {\it experiencer} of P, and observe that experience. The
SEM feature of ``$\phi_{exp}$ P-garu/gat-ta'' is the following.

\enumsentence{
\outerfs{rel:{\it garu}\\observer:$\:$\fbox{o} where \fbox{o} $\neq$ \fbox{e}\\
soa:\outerfs{rel:P\\exp:$\:$\fbox{e}\\}\\}\\

where `` $\neq$ '' means ``not token
identical.''}

In our FS, constraints for tokens like \fbox{o} are written with
``where'' as shown in this FS. Since constraint satisfaction method in
UG has been and is developed by many researchers recently i.e.
\cite{hasida}, our theory will be able to be implemented in systems
like theirs.

If the sentence finishes just after ``garu/gat-ta'', the important
points are 1) an introduced {\it observer} is the speaker, and
consequently 2) the {\it experiencer} cannot be the speaker. If a clause
with ``garu/gat-ta''is a subordinate clause, the {\it experiencer}
cannot be identified with a semantic role corresponding to the subject
of main clause or higher clause.

As for category 2, subjective verbs like ``kurusimu''(feel sick) and
``kanasimu''(feel sadness) that describe subjective and/or emotional
experience in verb form, are used.  Like the case of {\it garu}, an
{\it observer} who observers the experience can be introduced.  However this
observer is not obligatory. Therefore unlike the ``garu/gat-ta'' case,
the {\it experiencer} also can be an obligatory semantic role of higher
clause as well as the speaker.

\section{Complex Sentence}

\subsection{Feature Structure}

According to the hierachical structure of Japanese sentence shown in
Fig.1 , the essential part of hierarchical structure of the
following sentence (\ref{samuex}) is shown in Fig.2 .  In this
figure, the structure just below each proposition is replaced with the
corresponding parts of sentence.

\enumsentence{\label{samuex}
\begin{tabular}{lll}
$[\phi_{exp}$ & samu & -gat-ta\\
$[$ & feel cold & behaved like \\
\end{tabular}
\begin{tabular}{rlll}
node ] & $\phi_{agt}$ & mado -o & sime-ta.\\
because ] &  & window -ACC & closed.\\
\end{tabular}\\
`Since $\phi_{exp}$ behaved like feeling cold,
$\phi_{agt}$ closed the window.'\\}

\begin{center}
\begin{verbatim}

           ComplexProposition
             /      \
   Sub-Clause        Proposotion
     /      \            |
 Comment    Conjunt  `mado o sime-ta'
    |           |
Proposition  `node'
    |
`samu-gat-ta'

\end{verbatim}
Figure.2 : Hierarchical structure of (\ref{samuex})\\
\end{center}

Basically the embedding structure of FS corresponds to the hierarchy
shown in the hierarchical structure Fig.1 . To grasp the
image of the relation between a hierarchical structure and the
corresponding FS, we show an example of FS of the above complex
sentence (\ref{samuex}) analyzed based on this hierarchical structure
in the following.  This FS is the result of the unification between
the FSs of subordinate clause and main clause, where the contents of
syntactic feature HEAD , namely
\fbox{syn} is omitted.\\

\outerfs{    {\sc morph}:$\:$`samu-gat-ta node,mado o sime-ta' \\
             {\sc head}:$\:$ \fbox{syn}\\
             {\sc sem}:$\:$\fbox{matrix-sem}$\wedge$\fbox{sub-sem}
                 $\wedge$ \fbox{o} $\neq$ \fbox{c}\\}\\

\fbox{matrix-sem}=\outerfs{
              rel: {\it sime} \\
               agent:$\:$\fbox{o} \\ object:window\\tense:past\\
              }\\

\fbox{sub-sem}=
\outerfs{rel: {\it node} \\
            motivated: $\:$\fbox{o} \\
            soa:\outerfs{rel:{\it garu}\\observer:\fbox{o} \\
                soa:\outerfs{rel: {\it samu-i} \\
                experiencer:$\:$\fbox{c} \\
                tense:past \\ } \\}\\}\\

where English grosses of relation name is the following: {\it sime}:`close',
{\it node}:`because', {\it samu-i}:`feel cold'.

The key point of the semantics of complex sentence is the role {\bf
motivated} that appears in \fbox{sub-sem} which corresponds to the
content of the subordinate clause. The role {\it motivated} is the link
between the content of subordinate clause and the main clause.
Semantically {\it motivated} is characterized as the following.

\begin{definition}[Motivated]
{\it Motivated} is a person who is affected by the situation described
by the subordinate clause deeply enough to feel or act as the main
clause describes.
\end{definition}

The important and indispensable part of semantics of complex sentence
is, roughly speaking, the relation between a subordinate clause and
the main clause. But if you look more closely, this relation is
actually the relations among semantic/pragmatic roles appearing in the
subordinate clause and those appearing in the main clause. The newly
introduced role of {\it motivated} gives the most important clue for
this relation. Therefore, in the rest of this paper, our effort will
be concentrated into whom a {\it motivated} refers to. More precisely,
in FS, our main concerns are which semantic role in the SEM
of subordinate clause the {\it motivated} can or cannot be unified
with, and which semantic role in the SEM of main clause the
{\it motivated} can or cannot be unified with.

\subsection{Constraints}

In this subsection, we propose the constraints on complex sentence.  For
this, at first we categorize the relations between subordinate clause
and main clause based on their semantics. They are divided up to many
types of complex sentence. We show the most important and typical types
in Table.\ref{Table2}, where SC and MC mean `subordinate clause' and
`main clause' respectively.  In this table, the first column is for a
name of sentence type, the second column indicates a rough meaning of
the relation between subordinate clause SC and main clause MC of complex
sentence, and the third column shows Japanese conjunctive particles used
to represent a type of complex sentence in the same row.

\begin{table}
\begin{tabular}{|l|l|l|}\hline
type & outline meaning of & Japanese conjuncts\\ & complex sentence  & \\
\hline
1 & SC causes MC & {\it node, kara}\\
\hline
2 & Although SC, MC & {\it noni, ga, keredomo,}\\ & &{\it  temo, i-te,}\\
& & {\it  i-tutu, i-nagara}\\
\hline
3 & If SC then MC & {\it to, nara, tara, reba}\\
\hline
4 & When/after/before  & {\it toki, ato,}\\ & etc SC, MC & {\it mae,} etc.\\
\hline
\end{tabular}
\vspace*{-2mm}
\caption[Table2]{Clause Adjuncts}\label{Table2}
\end{table}

Three VP adjuncts, {\it te, tutu,} and {\it nagara}, are
usually used to express events ocurring simultaneously. However, if they
are used with aspectual suffix {\it i} which means perfective, for
instance {\it i-nagara}, they are regarded as clause conjuncts and are to
be interpreted as `although'\cite{minami}.  We don't deal with type 4,
because a temporal adverbial clause just describes an event that occurs
before, simultaneously or after another event which is described by the
main clause.  Therefore generally we don't expect essential information
for relations among semantic roles appearing in adverbial or main clause
from this type of sentence.

Now we focus on type 1,2 and 3, where a {\it motivated} plays the key role
in the constraints. In Table.\ref{Table3} we show the constraints that
say which semantic/pragmatic role of subordinate clause can be a {\it
motivated}. Table.\ref{Table4} shows which semantic role of main clause
can be unified with the {\it motivated}.  In these tables, the first
column of the first row is for constraint names, the second column shows
a set of sentence types for which the constraints shown in the second
row apply.  The third column of Table.\ref{Table3} shows predicate
patterns of subordinate clause, and the third column of
Table.\ref{Table4} shows semantic categories of predicate of main
clause. For them, constraints written in the second row apply.  Note
that all of these constraints in Table.\ref{Table3} are local in a
subordinate clause, because both sides of = of constraints are roles of
subordinate clause.  In case of subjective adjective without {\it garu},
the constraint `{\it motivated} = {\it experiencer}' holds also for type
1 except for the case where directionally auxiliary verb ``yaru(give)'',
``kureru(be given)''  are used. Analysis for these cases is one of
our future problem.

As for Table.\ref{Table4}, $\mbox{state}^\ast$ is a state except for the
case that there exists a third party who is a {\it motivated} puts the
{\it experiencer} into that state. For instance, the {\it experiencer}
is permitted to do something by the {\it motivated}. Since in this kind
of case things are quite complicated, we omit it here because of the
limited space.  Constraints in Table.\ref{Table4} are also local in a
main clause because every semantic role that appeares in the righthand
side of the constraints is defined within the main clause. Needless to
say, the influence from a subordinate clause comes only via role {\it
motivated}.

\begin{table}
\begin{tabular}{|l|l|l|}\hline
name & type & predicate \\\hline
\multicolumn{3}{|c|}{constraint} \\
\hline\hline
S1 & 1,3 & subjective adjective + {\it garu}   \\
   &     & verb + {\it ta} + {\it garu}  \\
\hline \multicolumn{3}{|c|}{{\it motivated} = {\it observer}}\\\hline\hline
S2 &  2 & subjective adjective + {\it garu}  \\
   &     & verb + {\it ta} + {\it garu}  \\\cline{2-3}
   & 1,2,3   & subjective verb  \\
\hline
\multicolumn{3}{|c|}{{\it motivated} = {\it observer} $\vee$ {\it experiencer}}
\\\hline\hline
S3 & 1 & subjective adjective (without {\it garu})\\\hline
\multicolumn{3}{|c|}{{\it motivated} = {\it experiencer}}\\\hline\hline
S4 & 2,3 & subjective adjective (without {\it garu})\\\hline
\multicolumn{3}{|c|}{{\it motivated} = {\it experiencer} $\vee$ {\it
observer}}\\\hline\hline
S5 &  1,2,3 & intransitive passive \\\hline
\multicolumn{3}{|c|}{{\it motivated} = {\it affected}}\\\hline\hline
S6 &  1,2,3 & transitive passive \\\hline
\multicolumn{3}{|c|}{ \begin{tabular}{ll}
{\it motivated} & =  {\it affected} :if {\it affected} exists,\\
       & = {\it patient} : otherwise\\
\end{tabular}}\\
\hline
\end{tabular}\\where `name' means a name of each constraint.
\vspace*{-2mm}
\caption[Table3]{Constraints in Subordinate Clause}\label{Table3}
\end{table}

\begin{table}
\begin{tabular}{|l|l|l|}\hline
name & type &  predicate category\\
\hline
\multicolumn{3}{|c|}{constraints}\\
\hline\hline
M1 & 1,2,3  &  action \\\hline
\multicolumn{3}{|c|} {sub-clause:SEM:{\it motivated}  =  {\it agent}} \\
\hline\hline
M2 & 1,2,3  &  $\mbox{state}^\ast$ \\
\hline
\multicolumn{3}{|c|} {sub-clause:SEM:{\it motivated}  = {\it experiencer}} \\
\hline
\end{tabular}
\vspace*{-2mm}
\caption[Table4]{Constraints in Main Clause}\label{Table4}\end{table}

In the rest of this section we show the examples that exemplify these
constraints.
\footnote{
The examples shown below are a tip of iceberg we actually analyzed, of
course. We gather the data about native's intuitive interpretation
from more than twenty natives around authors.}

First, we take (\ref{samuex}) of type 1. The constraints to be applied
are S1 and M1 as you know from the contents of subordinate and main
clause. By combination of S1 and M1, zero {\it agent} of main
clause:$\phi_{agt}$ is the {\it observer} of the situation described by
the subordinate clause, where $\phi_{exp}$ behaved like feeling cold.
This interpretation coincides with native's intuition.

Look at the following pair of example.

\enumsentence{\label{kuru1}
\begin{tabular}{lllr}
$[\phi_{exp}$ & kurusi & -gat-ta & noni]\\
$[$ & feel bad & behaved & but]\\
\end{tabular}
\begin{tabular}{lll}
kekkyoku & $\phi_{agt}$\\
at last &  \\
kusuri - o & nom & anakat-ta.\\
medicine - ACC & drink & not-PAST.\\
\end{tabular}\\
`Although $\phi_{exp}$ behaved like feeling bad, $\phi_{agt}$ didn't take a
medicine at last.'}

\enumsentence{\label{kuru2}
\begin{tabular}{llll}
$[ \phi_{exp}$ & nokori & ta & -gat-ta\\
$[$ & stay & want & behaved like\\
\end{tabular}
\begin{tabular}{rllll}
noni] & kekkyoku & $\phi_{agt}$ & oi & dasi-ta\\
but ] & finally  &  & forced &  out.\\
\end{tabular}\\
`Although $\phi_{exp}$ wanted to stay, $\phi_{agt}$ finally forced him out.'}

In both of (\ref{kuru1}) and (\ref{kuru2}), the {\it motivated}s of
subordinate clause are constrained by S2, namely {\it motivated}s can be
either $\phi_{exp}$ or the {\it observer} of subordinate clause.
Constraint M1 says that in both cases, $\phi_{agt}$ is unified with the
{\it motivated}. Intuitively in (\ref{kuru1}), $\phi_{agt}$ is
$\phi_{exp}$. On the other hand in (\ref{kuru2}), $\phi_{agt}$ is the
{\it observer}. Both of these interpretations comply with constraints
S2, and M1.
\enumsentence{\label{atuk}
\begin{tabular}{llr}
$[\phi_{1exp}$ & atui & node $]$ \\
$[$ & be hot & because $]$ \\
\end{tabular}
\begin{tabular}{ll}
 $\phi_{2exp}$ & komaru.\\
 & be in trouble.\\
\end{tabular}\\
`Since it is hot, I am in trouble.'}

Intuitively $\phi_{1exp}$ corefer with $\phi_{2exp}$. This
interpretation is expected by constraint S3 and M2 that apply in this
case. As you know from these examples, our constraints are not strong
enough to identify the antecedent of $\phi_{agt}$ uniquely, but makes
safe interpretations.  Moreover disambiguation done by these
constraints is useful for further inference that will be done with
commonsense knowledge or with a special vocabulary like
`kekkyoku(finally)' used in (\ref{kuru2}).

In case of S5, namely intransitive passive or adversity passive, it is
well known, i.e.
\cite{gunji87} that there exists a person who is affected by the
situation described by the passive sentence. An example sentence is
the following.

\enumsentence{\label{tuma}
\begin{tabular}{lrll}
$[ \phi_{affect}$ & tuma - ni & sin & -are \\
$[$ & wife & be dead & -PASSIVE \\
 -ta & noni]\\
 -PAST & but]\\
\end{tabular}
\begin{tabular}{lll}
$\phi_{exp}$ & kanasimi - mo -si & nai.\\
 & show sadness & not.\\
\end{tabular}\\
`Although his wife had gone, $\phi_{exp}$ doesn't show a bit of sadness.}

The semantic role of this affected person , in (\ref{tuma}) zero
 role:$\phi_{affect}$ whose wife was dead, is an {\it affected}.
The intuitive interpretation that $\phi_{exp} = \phi_{affect} (= {\it
motivated})$, is expected by our constraints: S5 of Table.\ref{Table3}
and M1 of Table.\ref{Table4}. On the contrary, in case of S6, namely
transitive passive, generally we don't have an {\it affected}.
However in some context, a transitive passive form may require the
role {\it affected} which is inherent to adversity passive.
For instance,
\enumsentence{\label{atama}
\begin{tabular}{lll}
$\phi_{affect}$ & saihu - ga & nusum\\
 & wallet - SUBJ & steal \\
\end{tabular}
\begin{tabular}{ll}
-are & -ta\\
-PASSIVE & -PAST\\
\end{tabular}\\
`$\phi_{affect}$'s wallet was stolen.'}

In this case, a person whose wallet was stolen is not explicit but
regarded as an {\it affected}.  Another case having an {\it affected}
is that a relational noun is the subject of transitive passive. Then a
person who is in the relation expressed by the relational noun is
thought to be affected by that situation ,too.  Here we take `mother',
`father', `daughter', `son', `supervisor', and so forth as a
relational noun. A couple of example sentences are the following.

\enumsentence{\label{kobun1}
\begin{tabular}{llrl}
$[$ kobun  & -ga & yar & -are\\
$[$ henchman & -SUBJ & attack & -PASSIVE\\
-ta & node ] & \\
-PAST & because] & \\
\end{tabular}
\begin{tabular}{lll}
 $\phi_{agt}$  & sikaesi - ni & it-ta.\\
& retaliate & go- PAST\\
\end{tabular}\\
`Since his henchman was attacked, the boss retaliated.'}

\enumsentence{\label{kobun2}
\begin{tabular}{llll}
$[$ kobun  & -ga & yar & -are\\
$[$ henchman & -SUBJ & attack & -PASSIVE\\
\end{tabular}
\begin{tabular}{lr}
-ta & noni]\\
-PAST & but ]\\
\end{tabular}
\begin{tabular}{ll}
$\phi_{agt}$ & te-o komaneite-iru.\\
 & did nothing.\\
\end{tabular}\\
`Although his henchman was attacked, the boss didn't retaliate.'}

$\phi_{agt}$ who retaliated (\ref{kobun1}) (or didn't retaliate
(\ref{kobun2})) has a certain relation between the henchman who had
been attacked. For instance, $\phi_{agt}$ may be the boss of that
henchman. In (\ref{kobun1}), since constraint S6 of
Table.\ref{Table3} and M1 of Table.\ref{Table4} apply, $\phi_{agt}$ is
an {\it affected} of attacking event described in the subordinate
clause. This interpretation coincides with native's intuition.

In sum, with these constraints, a constraint satisfaction
process in UG based parsing can be done locally and consequently very
efficiently.  In other words, primarily a constraint satisfaction
process of a subordinate clause can be done within the analysis of
subordinate clause, and that of the main clause can be done within it
except for using {\it motivated} whose value has already been constrained
in the subordinate clause.

\section{Related Works and Conclusions}

One of the relevant researches to ours is JPSG that has been developed by
Gunji\cite{gunji87,gunji89} and is further studied by the ICOT working
group. Our focus is a more pragmatics oriented one than JPSG is.
Many Japanese linguists have already done the enormous amount of basic
observations and proposed linguistic theories about the phenomena we
deal with in this paper
\cite{mikami,Kuno73,Kuno78,Ohye,minami,takubo,Teramura84,Teramura90,Saito}.
Of course our research is based on their works and observations. In
\cite{Ohye}, it is said that if {\it garu} is used in a subordinate
clause, the subject of the main clause is not the {\it experiencer} of the
subordinate clause. In
\cite{Saito}, she says that 1) a cognizer that corresponds to our
{\it observer} is introduced if {\it garu} is used, and 2) if an {\it
observer} is introduced in the subordinate clause, the mentally
responsible person appearing in the main clause is identical with the
{\it observer}. In linguistic phenomena, these observations are
similar to the constraint we propose here. So what is new? The answer
is that: 1) We explicitly state the semantics of complex sentence as
the relations among semantic roles.  Namely, since we use semantic/pragmatic
roles instead of grammatical roles in constraints, our constraints can
account for zero anaphora in a sentence where the main clause is
passive where an {\it agent} or an {\it experiencer} is not
necessarily the subject, like the following example.
\enumsentence{\label{pass}
\begin{tabular}{lllll}
Taro & -wa & $[$ gakkou e & iku-no & -wo \\
 & -Topic & $[$ to school &  go-NOM & -ACC\\
\end{tabular}
\begin{tabular}{llr}
iya & -gat-ta & node $]$ \\
hate & behaved like & because $]$\\
\end{tabular}
\begin{tabular}{llll}
$\phi_{agt}, \phi_{pat}$ & okor& -are & -ta.\\
& scold & -PASSIVE & -PAST\\
\end{tabular}\\
`Since Taro behaved like hating to go to school, he was scolded.' }
where the intuitive reading is the following: $\phi_{pat}$, that is zero
subject, refers to Taro, and $\phi_{agt}$, that is not the zero subject,
refers to Taro's parents who are the {\it observer} and {\it motivated}
of the subordinate clause.  2) We formalize this theory in UG formalism,
even though the details are omitted due to the space limitation. 3) We
find that the constraints of complex sentences are actually local ones.
This localization of constraint was found by introducing new pragmatic
roles {\it observer} and {\it motivated}, and is extremely important for
efficiency of UG based parsing. This localization also makes the
proposed constraints be compositional ones, because in the case of
deeply embedded complex sentence to identify the referent of each {\it
motivated} that bridges between a subordinate clause and its main
clause, the constraints we proposed are resolved with computation
confined within each clause.

Analysis of case in which a directional auxiliary verb i.e.
`yaru',`kureru' is used is left as the future problem. Finally, we
implemented a Japanese language understanding system based on the
theory we state in this paper, but due to the space limitation we will
report the detail of implementation in other place in the near future.

\end{document}